%% file: kFFNN_final.tex
\documentclass{article}
\usepackage{spconf} 
\usepackage{amsmath}
\usepackage{color}
\usepackage{times}
\usepackage{epsfig}
\usepackage{amssymb}
\usepackage{latexsym}
\usepackage{graphicx}
\usepackage{multirow}
\usepackage{epstopdf}
\usepackage{tipa}
\usepackage{multicol}
\usepackage{subfig}
\usepackage{tikz}
\usepackage{algorithm}
\usepackage{algorithmic}
\usepackage{graphics}
\usepackage{tabu}

\usepackage{times}
\usepackage{graphicx} 

\usepackage{algorithm}
\usepackage{algorithmic}

\usepackage{hyperref}

\def\paper{paper}
\def\myin{4}
\def\hi{2}
\def\ou{1}
\def\ndatseq{3}
\def\ndatsam{2}
\def\dataset{MediaEval}
\def\mycomment#1{{\color{black}{#1}}}
\def\totvideo{\mycomment{7571}}
\def\nfeatures{\mycomment{21}}
\def\nfeature{g}
\def\feature{\vec{\nfeature}}

\def\ffnnw{{^{(ih)}w}}
\def\ffnnv{{^{(ho)}w}}
\def\rnnw{{^{(ih)}w}}
\def\rnnv{{^{(ho)}w}}
\def\rnnh{{^{(hh)}w}}
\def\error{\epsilon}
\def\lrate{\eta}

\begin{document} 
\title{k-FFNN: Using a priori knowledge in Feed-forward Neural Networks}
\makeatletter
\def\name#1{\gdef\@name{#1\\}}
\makeatother \name{{\em Sri Harsha Dumpala, Rupayan Chakraborty, Sunil Kumar Kopparapu}}
\address{TCS Innovation Labs-Mumbai, Thane West, India\\
\\ 
  {\small \tt \{d.harsha,rupayan.chakraborty,sunilkumar.kopparapu\}@tcs.com}}

\maketitle
\begin{abstract} 
Recurrent neural network (RNN) are being extensively used over feed-forward
neural networks (FFNN) because of their inherent capability to capture temporal 
relationships that exist in the sequential data
such as speech. This aspect of RNN is advantageous especially when
there is no a priori knowledge about the temporal correlations within
the data. However, RNNs require large amount of data to learn these temporal
correlations, limiting their advantage in low resource scenarios. It is
not immediately clear (a) how a priori temporal knowledge can be used in a FFNN architecture 
(b) how a FFNN performs when provided with this
knowledge about temporal correlations (assuming available) during
training. The objective of this paper is to explore k-FFNN, namely a FFNN architecture that can incorporate 
the a priori knowledge of the temporal relationships
within the data sequence during training and compare k-FFNN performance
with RNN in a low resource scenario. We evaluate the performance of k-FFNN and RNN by 
extensive experimentation on MediaEval 2016 audio data (“Emotional Impact of Movies” task). 
Experimental results show that the performance of k-FFNN is comparable to RNN, and in 
some scenarios k-FFNN performs better than RNN when temporal knowledge is injected into FFNN architecture.
The main contributions of this paper are (a) fusing
a priori knowledge into FFNN architecture to construct a k-FFNN and
(b) analyzing the performance of k-FFNN with respect to RNN for different size of training data.
\end{abstract}

\section{Introduction}
\label{sec:introduction}

Artificial neural networks are extensively used in all types of classification 
problems \cite{Kohonen19883}. Speech technologies and applications are not exception to that.
Subsequent advancements in the field of neural networks was adapted to build better 
speech processing systems \cite{6296526}. Recurrent neural network (RNN) is one such 
advancement in neural networks, which is being used extensively to solve 
various problems in speech processing \cite{6638947,graves2014towards}.
Among those, audio (speech) emotion recognition is one of the latest developments which 
is an integral part of Human Computer Interaction (HCI) system. 
RNN has been successfully applied in speech emotion recognition \cite{WeningerRMS16}.

\begin{figure*}[!ht]
\centerline{
\hfill
\includegraphics[width=0.24\textwidth,height=3.4cm]{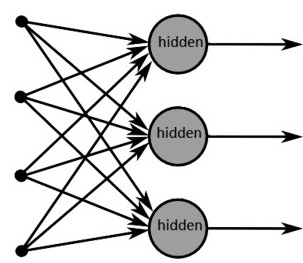}
\hfill
 \includegraphics[width=0.24\textwidth,height=3.4cm]{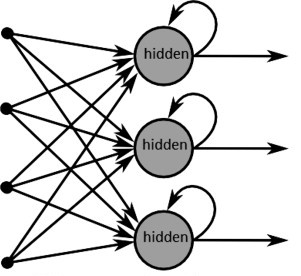}
\hfill
}
\centerline{\hfill (a) \hfill (b) \hfill}
\caption{{General structure of (a) feed-forward neural network and (b) recurrent neural network.}}
\label{fig:nn}
\end{figure*}

It is evident that a feed-forward neural network (FFNN) discriminatingly learns the patterns within the inputs, even from a 
low resource dataset \cite{Haykin}. But they are not designed to learn sequential relationships within the data. 
On the other side, deep neural networks like RNN has an inherent characteristic of learning and exploiting 
temporal relationships amongst 
the sequences \cite{Rumelhart,Werbos1988339,COGS}, however RNN requires large training data to capture those correlations.  
Figure \ref{fig:nn}(a) shows the general architecture of a FFNN and Figure
\ref{fig:nn}(b) shows the structure of a RNN. The essential difference between the two is the self 
loop within the hidden layer which is useful in capturing the unseen temporal relationship that might exist in the training data. 
The general structure of an FFNN as shown in Figure \ref{fig:nn}(a), consists of three layers i.e., 
input layer, hidden layer and output layer. The input data or features are fed to the input layer which pass through the hidden layer 
to the output layer. Here the input features or the posterior probabilities 
pass from the input layer to the hidden layer and then to the output layer but never in backward direction. 
Hence, the name feed-forward neural network. 
In a FFNN, a mapping is obtained between the input features and the output values in a supervised learning condition. 
By design in FFNN, no information regarding the sequence in which the inputs are fed to the 
network is captured. In FFNN, all the input data sequences are considered independent of 
each other. On the other hand a RNN network (see Figure \ref{fig:nn}(b))
is similar to FFNN except for the feedback loop in the hidden layer. 
This ensures the capture of temporal information in the sequence of inputs along with the mapping between the input and output is 
also captured. 

In this \paper\, we explore the use of knowledge regarding the temporal relationships within the sequence of training data 
while using a FFNN architecture for a limited resource scenario. 
Knowing that an RNN architecture is capable of inherently learning the temporal 
relationships that exist in the sequential data, we use RNN to automatically capture that information.
However, this aspect of RNN is useful especially when there is no a priori knowledge 
about the temporal correlations within the data. But to learn temporal correlations automatically, 
substantial amount of training data are required. 
\begin{quote}
\it What if a priori knowledge of the temporal relationships 
within the data sequences are known for limited samples, can a FFNN perform similar as an RNN?
\end{quote}
In this work, a FFNN architecture which can  use a priori knowledge of temporal correlationships 
has been explored, and we call it k-FFNN (short form of knowledge infused FFNN).
In particular, we capture the relationship between FFNN and RNN, and then subsequently show 
through extensive experiments that the knowledge of temporal relationship can be infused to
improve the performance of FFNN in a way that resembles RNN. 
Using MediaEval 2016 audio data (“Emotional Impact of Movies” task)\cite{MediaEval2016}), 
we conduct several experiments to establish that the performance of k-FFNN is comparable to RNN 
and in some scenarios k-FFNN outperform RNN. 
The main contributions of the paper are (a) incorporation of a priori temporal/sequential knowledge in FFNN to construct a k-FFNN and 
(b) experimentally showing that not only the performance of k-FFNN is as good as RNN, but also better when there is 
small amount of training data. The paper is organized as follows. 
Section \ref{sec:hypothesis} presents the hypothesis we make with some theoretical representations. 
In Section \ref{sec:dataset}, we discuss the dataset used to validate our hypothesis. Section \ref{sec:experiment}
describes the experiments conducted with an analysis. We conclude in Section \ref{sec:conclude}. 
\section{Hypothesis}
\label{sec:hypothesis}
We start off with the hypothesis 
\begin{quote}
\it FFNN infused with prior knowledge about temporal relationship between data is similar to
RNN in terms of performance
\end{quote}

As seen in Figure \ref{fig:nn} the primary difference between an RNN and a 
FFNN is the presence of the hidden layer feedback self loop in RNN 
which adds memory to the RNN network over time. The question that we are addressing is if a 
regular FFNN fed with the sequence based a priori knowledge (i.e. k-FFNN) can perform as well as an RNN. In other words, if we had
some a priori knowledge about the sequence can we use it without depending 
on RNN to learn it through its hidden layer feedback loop. This is very useful especially in the scenarios where the training 
data is sparse plus when we are aware of the sequential relationship between data a priori.
We validate the hypothesis that the performance of k-FFNN and RNN are similar through an extensive experimentation using the \dataset\
dataset.

\input rnn_kann_t_validation.tex

\setlength{\intextsep}{0pt}
\begin{algorithm}
\begin{algorithmic}
\STATE{Given: Input-output pairs (Table \ref{tab:rnn_in_out})}
\STATE{Given: The RNN configuration ($\myin-\hi-\ou$)}
\STATE{Given: Initialize weights; $[\rnnw]_{\myin \times \hi}$, $[\rnnv]_{\hi
\times \ou}$ and $[\rnnh]_{\hi \times \hi}$}
\STATE{}
\FOR{$l=1:\ndatsam$ pick the input-output pair ($\feature_{l,1},\feature_{l,2},
\feature_{l,3}, v_l$)}
\STATE{}
\FOR{$t=1:T$}
\STATE{}
\FOR{$k=1:\hi$} \STATE{Compute (\ref{eq:rnn_hidden}), namely,
$ h_k^{t} =  \frac{1}{1 + \exp^{-\lambda (\sum_{j=1}^{\myin} (\nfeature^j_{lt})
(\rnnw_{jk}) + \sum_{h'=1}^{\hi}(h_{h'}^{t-1})(\rnnh_{h'k}))}}$ }
\ENDFOR
\ENDFOR
\STATE{}
\STATE{Compute (\ref{eq:rnn_out}), namely, 
$ o_l = \frac{1}{1 + \exp^{-\lambda (\sum_{k=1}^{\hi} (h_{k}^{T}) (\rnnv_{k}) )}} $}
\STATE{}
\STATE{Compute the error (\ref{eq:rnn_error}), namely, 
$\error = (o_l-v_l)^2$}
\STATE{}
\STATE{Compute 
$ [\Delta \rnnv]_{\hi\times\ou}$,
$[\Delta \rnnw]_{\myin\times\hi}$ 
and 
$[\Delta \rnnh]_{\hi\times\hi}$
using (\ref{eq:rnn_wm_out}), (\ref{eq:rnn_wm_forward}) and (\ref{eq:rnn_wm_fb}) 
}
\STATE{}
\STATE{Update weights (\ref{eq:rnn_wp}), namely,}
\STATE{$ [\rnnv]_{\hi
\times \ou} \leftarrow  [\rnnv]_{\hi
\times \ou} +  [\Delta \rnnv]_{\hi
\times \ou}$ }
\STATE{$ [\rnnw]_{\myin
\times \hi} \leftarrow [\rnnw]_{\myin
\times \hi} + [\Delta \rnnw]_{\myin
\times \hi}$}
\STATE{$ [\rnnh]_{\hi
\times \hi} \leftarrow [\rnnh]_{\hi
\times \hi} + [\Delta \rnnh]_{\hi
\times \hi}$}
\ENDFOR
\STATE{}
$ [\rnnw]_{\myin
\times \hi}$, $ [\rnnv]_{\hi
\times \ou}, [\rnnh]_{\hi \times \hi}$ represent the RNN for the input data (Table
\ref{tab:rnn_in_out}).
\end{algorithmic}
\caption{RNN Training}
\label{algo:rnn_training}
\end{algorithm}
\setlength{\intextsep}{0pt}
\subsection{RNN}

In RNNs, length of the input sequence (here $T=3$) apart from the values of the input sequence at each time instant is considered
to train the networks. The output of the hidden layer in case of RNNs is given as

\begin{equation}
h_k^{t} =  \frac{1}{1 + \exp^{-\lambda (\sum_{j=1}^{\myin} (\nfeature^j_{1t})
(\rnnw_{jk}) + \sum_{h'=1}^{\hi}(h_{h'}^{t-1})(\rnnh_{h'k}))}}
\label{eq:rnn_hidden}
\end{equation}
The output of RNN is given by
\begin{equation}
 o_1 = \frac{1}{1 + \exp^{-\lambda (\sum_{k=1}^{\hi} (h_{k}^{T}) (\rnnv_{k}) )}}
 \label{eq:rnn_out}
\end{equation}

The error in the output estimation is
\begin{equation}
 \error = (o_1 - v_1)^2
 \label{eq:rnn_error}
\end{equation}

The error is backpropagated through the length of the sequence i.e., back propagation through time (BPTT) is used to modify
the weights of RNN.

The weight modification of RNNs in general is

\begin{equation}
 \Delta \rnnw_{ij} = \Delta \rnnv_{ij} = \sum_{t=1}^{T=3}\delta_{j}^{t}a_{i}^{t}
 \label{eq:rnn_wm}
\end{equation}
and
\begin{equation}
 \Delta \rnnh_{ij} = \sum_{t=1}^{T=3}\delta_{j}^{t}a_{i}^{t-1}
 \label{eq:rnn_hwm}
\end{equation}
where $a_{i}^{t}$ is the activation function at the $i^{th}$ unit. So the weight modification for the output layer units is:
\begin{equation}
 \Delta \rnnv_{jk} = \sum_{t=1}^{T}\delta_{k}^{t}h_{j}^{t}
\end{equation}

where $\delta_{k}^{t} = (o^t-v^t)$

In the network architecture considered for this analysis, the output is available only at $ t=T=3$ is an example. The weight modification 
at output layer gets modified as

\begin{equation}
 \Delta \rnnv_{jk} = \lrate (o_1 - v_1). h_{j}^{T}
 \label{eq:rnn_wm_out}
\end{equation}

The modification of the input to hidden layer weights is obtained as

\begin{equation}
 \Delta \rnnw_{ij} = \sum_{t = 1}^{T}\delta_{j}^{t}\nfeature^{i}_{1t}
 \label{eq:rnn_wm_forward}
\end{equation}

and the recursive/hidden weights are modified as

\begin{equation}
 \Delta \rnnh_{ij} = \sum_{t = 2}^{T}\delta_{j}^{t} h_{i}^{t-1}
 \label{eq:rnn_wm_fb}
\end{equation}

here sigmoid activation function is considered for the hidden layers where $\delta_{k}^{t}$ is generally defined as

\begin{equation}
 \delta_{j}^{t} = h_{j}^{t}(1-h_{j}^{t})\left((\sum_{k = 1}^{\ou}\delta_{k}^{t}\rnnv_{jk} + \sum_{h' = 1}^{\hi}\delta_{h'}^{t+1}
 \rnnh_{jh'}\right)
\end{equation}

For the network architecture considered, the above equation gets modified as
\begin{equation}
 \delta_{j}^{T} = h_{j}^{T}(1-h_{j}^{T})(o_1-v_1)\rnnv_{j1}
\end{equation}
and
\begin{equation}
 \delta_{j}^{t} = h_{j}^{t}(1-h_{j}^{t})\left(\sum_{h' = 1}^{\hi}\delta_{h'}^{t+1}\rnnh_{jh'}\right)
\end{equation}

So the weights are modified as
\begin{eqnarray}
 \rnnv_{jk} \leftarrow  \rnnv_{jk} +  \Delta \rnnv_{jk}  \nonumber \\
	 \rnnw_{ij} \leftarrow \rnnw_{ij} + \Delta \rnnw_{ij} \nonumber \\
	 \rnnh_{ij} \leftarrow \rnnh_{ij} + \Delta \rnnh_{ij}
\label{eq:rnn_wp}
\end{eqnarray}

RNN is trained using data shown in Table \ref{tab:rnn_in_out}. Note that there is no $f()$ in the output and there is an
additional weights that represents the RNN (see Table \ref{tab:diff}). The training process is shown in Algorithm \ref{algo:rnn_training}.

\input{data_preperation.tex}

\section{Experimental Validation}
\label{sec:experiment}
In all our experiments, we used the audio extracted from $\totvideo$ video clips
(\dataset\ database) each of around $n$ ($n=8-10$)
seconds duration \cite{ProceedingsMediaeval16}. The database has a valence (and arousal) value in the range $[0,5]$ 
for all the  $\totvideo$ videos, namely $(c_k;v_k)$ for $k=1, 2, \cdots,
\totvideo$ is available. We
constructed   $c_{k1}, c_{k2}, \cdots, c_{kn}$ each of $1$ second duration 
from $c_k$ of $n$ second duration for $k=1, 2, \cdots, \totvideo$ 
(see (\ref{eq:concatenation})). For each
$\{c_{kj}\}_{k=1,j=1}^{k=\totvideo, j=n}$  we extracted 384 features, which were used for Interspeech 2009
Emotion Challenge \cite{SchullerSB09} using the openSMILE toolkit \cite{openSMILE}. We used WEKA Toolkit \cite{WEKA} to reduce
the feature dimension to $21$ using feature selection method. 

If $\feature_{k,j}$ represent the extracted features from $c_{kj}$ then 
the dataset used in our experiments for RNN training is shown in Table \ref{tab:rnnd}, and for FFNN 
set of experiments we constructed $v_{ki}$ as
mentioned in (\ref{eq:in_out}) resulting in a dataset as shown in Table \ref{tab:ffnnd}. 
We used a variety of $f(i)$'s to capture fade-in and fade-out in our experiments as shown in
Table \ref{tab:fis}.

\begin{table}
\begin{center}
\scalebox{0.9}{
\begin{tabular}{|llll|c|} \hline
\multicolumn{4}{|c|}{Input} & Output \\ \hline
\hline
$\feature_{1,1}$ & $\feature_{1,2}$ & $\cdots$&  $\feature_{1,n}$ & $v_1$ \\
$\feature_{2,1}$ & $\feature_{2,2}$ & $\cdots$&  $\feature_{2,n}$ & $v_2$ \\
$\cdots$ & $\cdots$ & $\cdots$ & $\cdots$ & $\cdots$ \\
$\feature_{k,1}$ & $\feature_{k,2}$ & $\cdots$&  $\feature_{k,n}$ & $v_k$ \\
$\cdots$ & $\cdots$ & $\cdots$ & $\cdots$ & $\cdots$ \\
$\feature_{\nfeatures, 1}$ & $\feature_{\nfeatures, 2}$ & $\cdots$&
$\feature_{\nfeatures, n}$ & $v_{\nfeatures}$ \\ \hline
\end{tabular}
}
\end{center}
\caption{Dataset used in our experiments for RNN.}
\label{tab:rnnd}
\end{table}

\begin{table}
\begin{center}
\resizebox{\columnwidth}{!}{%
\begin{tabular}{|c|c|c|c|c|c|c|} \hline
Input & Output & Input & Output& & Input & Output \\ \hline
\hline
$\feature_{1,1}$ & $f(1)v_1$ & 
$\feature_{1,2}$ & $f(2)v_1$ & 
$\cdots$&  
$\feature_{2,n}$ & $f(n)v_1$ \\
$\feature_{2,1}$ & $f(1)v_2$ & 
$\feature_{2,2}$ & $f(2)v_2$ & 
$\cdots$&  
$\feature_{2,n}$ & $f(n)v_2$ \\
$\cdots$ & $\cdots$ & 
$\cdots$ & $\cdots$ & 
$\cdots$&  
$\cdots$ & $\cdots$ \\
$\feature_{k,1}$ & $f(1)v_k$ & 
$\feature_{k,2}$ & $f(2)v_k$ & 
$\cdots$&  
$\feature_{k,n}$ & $f(n)v_k$ \\
$\cdots$ & $\cdots$ & 
$\cdots$ & $\cdots$ & 
$\cdots$&  
$\cdots$ & $\cdots$ \\
$\feature_{\nfeatures,1}$ & $f(1)v_{\nfeatures}$ & 
$\feature_{\nfeatures,2}$ & $f(2)v_{\nfeatures}$ & 
$\cdots$&  
$\feature_{\nfeatures,n}$ & $f(n)v_{\nfeatures}$ \\ \hline
\end{tabular}
}%
\end{center}
\caption{Dataset used in our experiments for k-FFNN.}
\label{tab:ffnnd}
\end{table}

\begin{table}
\centering
\begin{center}
\resizebox{\columnwidth}{!}{%
\begin{tabular}{|cccccccccc|l|} \hline
$f(1)$&$f(2)$&$f(3)$&$f(4)$&-&-&-&-&$f(n-1)$&$f(n)$ & Type
\\ \hline 
\hline
1&1&1&1&1&1&1&1&1&1 & FFNN \\ \hline
0.75& 0.9& 1& 1& 1& 1& 1& 1& 0.9& 0.75 & Fn1 \\ \hline
0.3& 0.6& 1& 1& 1& 1& 1& 1& 0.6& 0.3 & Fn2 \\ \hline
0.1& 0.2& 1& 1& 1& 1& 1& 1& 0.2& 0.1 & Fn3 \\ \hline
\end{tabular}%
}
\caption{Different $f(i)$ used in our experiments.}
\label{tab:fis}
\end{center}
\end{table}

\subsection{Experimental Analysis}

\begin{figure}
\centerline{\includegraphics[width=0.4\textwidth,height=3cm]{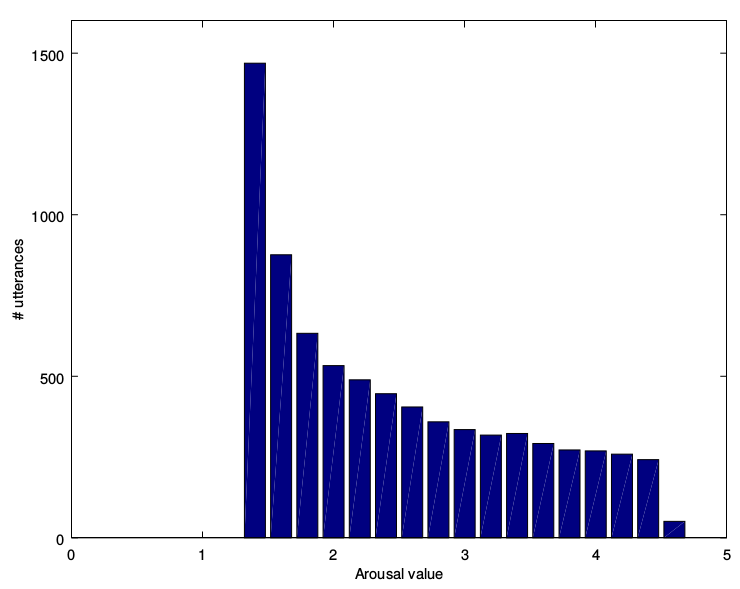}}
\caption{Histogram of arousal values}
\label{Histograms}
\end{figure}

The performance of the proposed k-FFNN system is compared with the popular RNN architecture i.e., simple RNN,
RNN with Long-short-term-memory (LSTM) units and bi-directional RNN with LSTM (BLSTM) units).
In this analysis, all k-FFNN and RNN systems are implemented using Keras deep learning toolkit \cite{KERAS}.
The architectures of the systems considered in this analysis are shown in Table \ref{tableArch}.
For all systems, only a single hidden layer is considered. The hidden layer size is selected by varying the number of units from 11 
(half of the sum of number of input (i.e., 21) and output units (i.e., 1)) to 44 (twice the sum of number of input and output units)
and selecting the number of nodes in the hidden layer which results in the best performance. Sigmoid (S in Table 
\ref{tableArch}) is used as the  
non-linear activation function on
the hidden units. The input layer has 21 linear (L) units and the output layer has a single linear (L) unit. 

\begin{table}
\begin{center}
\begin{tabular}{|l|c|c|} \hline
Model Name & Architecture \\ \hline
\hline
k-FFNN (Fn1) & 21L 21S 1L \\ \hline
  k-FFNN (Fn2) & 21L 21S 1L \\ \hline
  k-FFNN (Fn3) & 21L 21S 1L \\ \hline
  FFNN & 21L 22S 1L \\ \hline
  RNN & 21L 21S 1L \\ \hline
  LSTM & 21L 21S 1L \\ \hline
  BLSTM & 21L 21S 1L \\ \hline
\end{tabular}
\end{center}
\caption{System architecture details}
\label{tableArch}
\end{table}
The system performance is evaluated in terms of Mean Squared Error (MSE) and Pearson Correlation Coefficient (PCC).
PCC along with MSE is used as a performance metric as PCC provides a better evaluation of the performance of the
systems trained on datasets with output values arousal) distributed as shown in Figure \ref{Histograms}. It can be observed from 
Figure \ref{Histograms} that the
output values for arousal are concentrated more at a single value (at 1.5) compared to other values.
For the considered metrics, lower the MSE values better is the performance of the system and higher the PCC values, better is the
performance of the system.

The MSE and PCC values are computed at audio clip level for all the systems (k-FFNN and RNNs) to evaluate the performance. 
In case of RNNs, single output value $v_k$
is obtained for the given audio clip ($c_k$) containing the sequence $c_{k1}, c_{k2}, \cdots, c_{kn}$. 
Subsequently, the computation of MSE and PCC is
straight forward in case of RNNs. In case of a k-FFNN system (as shown in Table \ref{tab:ffnnd}), for each subsegment $c_{k1}, c_{k2},
\cdots, c_{kn}$ corresponding to the audio clip $c_k$, arousal/valence value are generated. 
To compute the MSE and PCC values for each audio clips $c_{k}$, the output
values obtained for each clips are scaled with a value depending on the function selected during training. Then
the mean of the values obtained at all subsegments is computed and compared with the original value $v_k$ assigned to that audio clip to
obtain the MSE and PCC values. If $v_1^{'}$, $v_2^{'}$, $v_3^{'}$, $...$, $v_n^{'}$ are the output obtained for all the audio 
segment corresponding to the audio clip $c_{k}$, then 
\begin{equation}
 V^{'} = \sum_{i = 1}^{n}v_{i}^{'}(1/f(i))
 \end{equation}
 is the defined arousal/valence value of the audio clip $c_{k}$. 

\begin{figure}
\centerline{\includegraphics[width=0.5\textwidth]{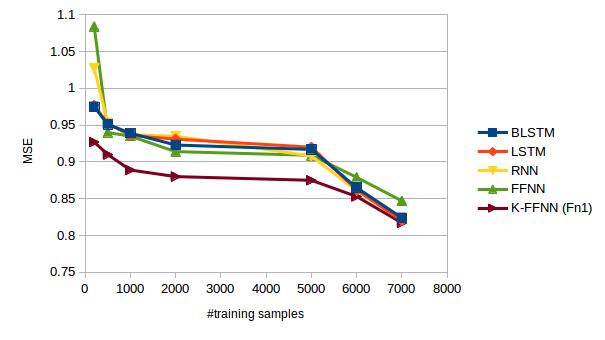}}
\caption{MSE values across different training set sizes}
\label{MSE values}
\end{figure}

\begin{figure}
\centerline{\includegraphics[width=0.5\textwidth]{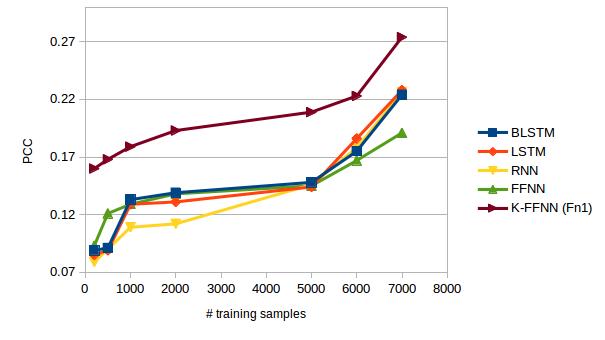}}
\caption{PCC values across different training set sizes}
\label{PCC values}
\end{figure}

The MSE and PCC values obtained by considering training sets of different sizes are shown in Figure \ref{MSE values} and Figure \ref{PCC values}, 
respectively for
different systems. It can be observed from Figure \ref{MSE values} that the MSE values obtained for k-FFNN (Fn1) is always lower or
equal to that of the MSE values obtained for RNNs for all sizes of training set. k-FFNN system performs much better than RNN 
systems for smaller training dataset. It can be observed that there is a performance improvement of 0.05 (MSE) when 200 training 
samples are used. Note that the MSE of RNN is 0.977 compared to MSE of 0.927 for k-FFNN (for 200 training samples) with function Fn1.
However the performance of k-FFNN closer to that of RNN when 6814 (90\% of dataset) samples are used for training. 
The MSE values are lower for FFNN compared to RNN for smaller training sets but are higher when 
the training set size is increased (MSE is 0.940 for FFNN and 0.953 for RNN when 500 samples are considered and 0.847 for FFNN and 0.820
for RNN when 6814 samples are considered). This shows that the performance of k-FFNN in terms of MSE is better than FFNN and RNN,
especially for smaller training set.  


It can be observed from Figure \ref{PCC values} that the PCC values are consistently higher for k-FFNN when compared to RNNs. Similar to
MSE, the variation in PCC values between k-FFNN and RNN is larger for smaller training sets (difference = 0.08 (0.079 for RNN and
0.16 for Fn1) when 200 train samples are considered) and gradually decreases when the size of the training set is increased
(difference = 0.048 (0.226 for RNN and 0.274 for k-FFNN), when 6814 train samples are considered). The PCC values obtained for FFNN are
lower compared to k-FFNN for all sizes of training set. The PCC values are higher for FFNN compared to RNNs for smaller training sets
but are lower when size of the training set is increased (PCC is 0.093 for FFNN and 0.079 for RNN when 200 samples are considered and
0.191 for FFNN and 0.226 for RNN when 6814 samples are considered).

Table 10 shows the MSE and PCC values obtained on arousal values by considering different knowledge infused functions (shown in Table \ref{tab:fis}) to represent
the temporal information. It can be observed from Table \ref{table 10} that the performance of the k-FFNN system trained by considering Fn1
performs better than the systems trained using Fn2 and Fn3 (both in terms of MSE and PCC). It is to be observed that the performance of
the k-FFNN systems developed by considering Fn2 and Fn3 is lower than FFNN. This shows that choosing a proper function (knowledge of sequential 
temporal correlations between data) to represent
the temporal information is critical for the performance of the k-FFNN and any arbitrary function used to represent the temporal
information will not improve the performance of k-FFNN but may even degrade the performance.

\begin{table}
\begin{center}
\begin{tabular}{|l|c|c|} \hline
Function & MSE & PCC\\ \hline
\hline
Fn1 & 0.820 & 0.274 \\ \hline
Fn2 & 0.871 & 0.185 \\ \hline
Fn3 & 1.55 & 0.059 \\ \hline
\end{tabular}
\end{center}
\caption{Performance evaluation of different k-FFNN systems.}
\label{table 10}
\end{table}

Table \ref{table 11} shows the MSE and PCC values obtained for the task of estimating the valence values for different systems. 
The MSE and PCC values are listed for systems trained on 6814 utterances. It can be observed that the performance of k-FFNN system
(using Fn1) is better than RNN and FFNN systems. But the performance of k-FFNN systems (using Fn2 and Fn3) are lower than RNN and even
FFNN. Hence the observations made from the prediction of arousal values is further supported by the results obtained for prediction
of valence values.

\begin{table}
\begin{center}
\begin{tabular}{|l|c|c|} \hline
System & MSE & PCC\\ \hline
\hline
k-FFNN (Fn1) & 0.319 & 0.128 \\ \hline
k-FFNN (Fn2) & 0.454 & 0.029 \\ \hline
k-FFNN (Fn3) & 0.762 & -0.051 \\ \hline
RNN & 0.331 & 0.126 \\ \hline
LSTM & 0.327 & 0.124 \\ \hline
BLSTM & 0.329 & 0.122 \\ \hline
FFNN & 0.343 & 0.106 \\ \hline
\end{tabular}
\end{center}
\caption{Performance evaluation of different systems for prediction of valence for training dataset of 6814 samples.}
\label{table 11}
\end{table}

\input conclusions.tex

\newpage
\bibliographystyle{IEEEbib}
\bibliography{kFFNN_final.bib}
\end{document}

%% file: rnn_kann_t_validation.tex
\begin{figure}
\centerline{\input{dia_ffnn_421.tex}}
\caption{FFNN $\myin-\hi-\ou$}
\label{fig:ffnn421}
\end{figure}
\setlength{\intextsep}{0pt}
\subsection{Background}
We validate our hypothesis by considering a simple network configuration
and derive a set
of expressions that show the relationship between k-FFNN and RNN. For sake of
simplicity, we consider a $\myin-\hi-\ou$ (input-hidden-output nodes)
network configuration. Additionally,
we consider a data sequence of length \ndatseq. Consider the data shown in Table
\ref{tab:data}.
\begin{table}
\begin{center}
\begin{tabular}{|lll|c|} \hline
\multicolumn{3}{|c|}{Input} & Output \\ \hline
\hline
$\feature_{1,1}$ & $\feature_{1,2}$ & $\feature_{1,3}$&   $v_1$ \\
$\feature_{2,1}$ & $\feature_{2,2}$ & $\feature_{2,3}$&  $v_2$ \\ \hline
\end{tabular}
\end{center}
\caption{Sample input-output training data.}
\label{tab:data}
\end{table}
More elaborately, if each $\feature_{i,j}$ was of dimension $\myin$ then we
would have Table \ref{tab:rnn_in_out} used as the input-output data to train
RNN. 
We assume that there exists some temporal relationship between $\feature_{1,1}$, $\feature_{1,2}$, $\feature_{1,3}$ 
and $\feature_{2,1}$, $\feature_{2,2}$, $\feature_{2,3}$, which can be captured as shown in Table \ref{tab:ffnn_in_out} . 
Namely, the output $v_1$ associated with $\feature_{1,1}$, $\feature_{1,2}$, $\feature_{1,3}$ is actually $f(1) v_1$, $f(2) v_1$, 
$f(3) v_1$ instead of $v_1$, $v_1$, $v_1$. This $f()$ is the mode in which the a priori temporal knowledge existing between 
 $\feature_{1,1}$, $\feature_{1,2}$, $\feature_{1,3}$ is infused into the training set. Note that a FFNN that uses training data
 as shown in Table \ref{tab:ffnn_in_out} is what we call k-FFNN.
 \begin{table}
\begin{center}
\begin{tabular}{|llll|c|} \hline
\multicolumn{4}{|c|}{Input} & Output \\ \hline
\hline
$\nfeature^1_{1,1}$ & $\nfeature^2_{1,1}$ & $\nfeature^3_{1,1}$&
$\nfeature^4_{1,1}$ & - \\
$\nfeature^1_{1,2}$ & $\nfeature^2_{1,2}$ & $\nfeature^3_{1,2}$&
$\nfeature^4_{1,2}$ & - \\
$\nfeature^1_{1,3}$ & $\nfeature^2_{1,3}$ & $\nfeature^3_{1,3}$&
$\nfeature^4_{1,3}$ & $v_1$ \\ \hline
$\nfeature^1_{2,1}$ & $\nfeature^2_{2,1}$ & $\nfeature^3_{2,1}$&
$\nfeature^4_{2,1}$ & - \\
$\nfeature^1_{2,2}$ & $\nfeature^2_{2,2}$ & $\nfeature^3_{2,2}$&
$\nfeature^4_{2,2}$ & - \\
$\nfeature^1_{2,3}$ & $\nfeature^2_{2,3}$ & $\nfeature^3_{2,3}$&
$\nfeature^4_{2,3}$ & $v_2$ \\ \hline
\end{tabular}
\end{center}
\caption{Input-output pair for training RNN.}
\label{tab:rnn_in_out}
\end{table}
\begin{table}
\begin{center}
\begin{tabular}{|llll|c|} \hline
\multicolumn{4}{|c|}{Input} & Output \\ \hline
\hline
$\nfeature^1_{1,1}$ & $\nfeature^2_{1,1}$ & $\nfeature^3_{1,1}$&
$\nfeature^4_{1,1}$ & $f(1) v_1$ \\
$\nfeature^1_{1,2}$ & $\nfeature^2_{1,2}$ & $\nfeature^3_{1,2}$&
$\nfeature^4_{1,2}$ & $f(2) v_1$ \\
$\nfeature^1_{1,3}$ & $\nfeature^2_{1,3}$ & $\nfeature^3_{1,3}$&
$\nfeature^4_{1,3}$ & $f(3) v_1$ \\ \hline
$\nfeature^1_{2,1}$ & $\nfeature^2_{2,1}$ & $\nfeature^3_{2,1}$&
$\nfeature^4_{2,1}$ & $f(1) v_2$ \\
$\nfeature^1_{2,2}$ & $\nfeature^2_{2,2}$ & $\nfeature^3_{2,2}$&
$\nfeature^4_{2,2}$ &  $f(2) v_2$\\
$\nfeature^1_{2,3}$ & $\nfeature^2_{2,3}$ & $\nfeature^3_{2,3}$&
$\nfeature^4_{2,3}$ & $ f(3) v_2$ \\ \hline
\end{tabular}
\end{center}
\caption{Input-output pair for training k-FFNN.}
\label{tab:ffnn_in_out}
\end{table}

For a 
 $\myin-\hi-\ou$ k-FFNN configuration, the model would be represented by
a total of $(\myin \times \hi) + (\hi \times \ou) = 10$ variables that
represent the network. Namely,
$[\ffnnw_{ij}]_{\myin \times \hi}$ the weights connecting the input to the hidden
layer and $[\ffnnv_{ij}]_{\hi \times \ou}$.
While in case of RNN the model
consists of not only $[\rnnw_{ij}]_{\myin \times \hi}$ the weights connecting
the input to the hidden
layer and $[\rnnv_{ij}]_{\hi \times \ou}$ the weights connecting the hidden and
the output layer but also $[\rnnh_{ij}]_{\hi \times \hi}$ the feedback connection
between the hidden layers. So in case of RNN, it is modeled by $(\myin \times
\hi) + (\hi \times \hi) + (\hi \times \ou) = 14$ variable. The input data remaining
the same, the differences in k-FFNN and RNN is captured in Table \ref{tab:diff}.

\begin{table}
\begin{center}
\begin{tabular}{|l|c|c|} \hline
Label & k-FFNN ($\myin-\hi-\ou$)& RNN ($\myin-\hi-\ou$)\\ \hline
\hline
Weights & - & $[\rnnh_{ij}]_{\hi \times \hi}$ \\
Output &$f(1), f(2), f(3)$  & - \\ \hline
\end{tabular}
\end{center}
\caption{Differences in terms of model and data between k-FFNN and RNN}
\label{tab:diff}
\end{table}

Assuming the same initial weights for both k-FFNN and RNN, we elaborate the
process of how the weights (or the model) gets updated as it is trained. We
assume the usual back-propagation based weights update. 
\setlength{\intextsep}{0pt}
\begin{algorithm}
\begin{algorithmic}
\STATE{Given: Input-output pairs (Table \ref{tab:ffnn_in_out})}
\STATE{Given: The k-FFNN configuration ($\myin-\hi-\ou$)}
\STATE{Given: The initial random weights; $[\ffnnw]_{\myin \times \hi}$, $[\ffnnv]_{\hi
\times \ou}$}
\STATE{}
\FOR{$i=1:\ndatseq, l=1:\ndatsam$ pick the input-output pair ($\feature_{li},f(i)v_l$)}
\STATE{}
\FOR{$k=1:\hi$} \STATE{Compute (\ref{eq:out_hidden}), namely, 
$ h_k = \frac{1}{1+ \exp ^{-\lambda (\sum_{j=1}^{\myin} \nfeature^j_{il}
\ffnnw_{jk} )}}$ }
\ENDFOR
\STATE{}
\STATE{Compute (\ref{eq:out_out}), namely, 
$o_i = \frac{1}{1+ \exp ^{-\lambda (\sum_{k=1}^{\hi} h_{k} \ffnnv_{k}
)}} $}
\STATE{}
\STATE{Compute the error (\ref{eq:ffnn_error}), namely, 
$\error = (o_i - f(i) v_l)^2$}
\STATE{}
\STATE{Compute 
$ [\Delta \ffnnv]_{\hi\times\ou}$
and
$[\Delta \ffnnw]_{\myin\times\hi}$ 
using (\ref{eq:ffnnw_delta}) and (\ref{eq:ffnnv_delta})
}
\STATE{}
\STATE{Update weights (\ref{eq:ffnn_wp}), namely,}
\STATE{$ [\ffnnv]_{\hi
\times \ou} \leftarrow  [\ffnnv]_{\hi
\times \ou} +  [\Delta \ffnnv]_{\hi
\times \ou}$ }
\STATE{$ [\ffnnw]_{\myin
\times \hi} \leftarrow [\ffnnw]_{\myin
\times \hi} + [\Delta \ffnnw]_{\myin
\times \hi}$}
\ENDFOR
\STATE{}
$ [\ffnnw]_{\myin
\times \hi}$, $ [\ffnnv]_{\hi
\times \ou}$ represent the k-FFNN for the input data (Table
\ref{tab:ffnn_in_out}).
\end{algorithmic}
\caption{FFNN Training}
\label{algo:ffnn_training}
\end{algorithm}
In case of FFNN (see
Figure \ref{fig:ffnn421}), the
output of the hidden node is given by
\begin{equation}
h_k = \frac{1}{1+ \exp ^{-\lambda (\sum_{j=1}^{\myin} \nfeature^j_{11}
\ffnnw_{jk} )}}  
\label{eq:out_hidden}
\end{equation}
assuming the sigmoid to be the squashing transfer function and
$\feature_{11}$ is the input and $\lambda$ is a constant which determines the
steepness of the sigmoid.
Similarly the output would be
\begin{equation}
o_1 = \frac{1}{1+ \exp ^{-\lambda (\sum_{k=1}^{\hi} h_{k} \ffnnv_{k} )}}  
\label{eq:out_out}
\end{equation}
Now the error 
\begin{equation} 
\error = (o_1 - f(1) v_1)^2
\label{eq:ffnn_error}
\end{equation} is used to modify the  weights ($\ffnnw,
\ffnnv$) such that when the same input $\feature_{11}$ is given to k-FFNN it
would reduce $\error$ (called back propagation of error) generally using the steepest descent algorithm.
The weight, $\ffnnv_{k}$ for $k=1,2$ and the weight, $\ffnnw_{jk}$ for $j = 1, 2, 3, 4; k=1,2$,
for the hidden layer are modified as 
\begin{eqnarray}
 \ffnnv_{k} \leftarrow  \ffnnv_{k} +  \Delta \ffnnv_{k}  \nonumber \\
	 \ffnnw_{jk} \leftarrow \ffnnw_{jk} + \Delta \ffnnw_{jk}
\label{eq:ffnn_wp}
\end{eqnarray}
where 
\begin{equation}
 \Delta \ffnnv_{k} = \lrate (o_1 - f(1) v_1). h_{k} 
\label{eq:ffnnv_delta}
\end{equation}
\begin{equation}
 \Delta \ffnnw_{jk} = \lrate (o_1 - f(1) v_1). h_{k} (1 - h_{k}) \ffnnw_{jk} \nfeature^j_{11}
\label{eq:ffnnw_delta}
\end{equation}
The next input, namely $\feature_{12}$, is passed
through the network to obtain $h_k$ (\ref{eq:out_hidden}) and $o_1$
(\ref{eq:out_out}). Now $o_1$ is used to compute the error
(\ref{eq:ffnn_error}) followed by weight update (\ref{eq:ffnn_wp}). This
continues for other inputs (Table \ref{tab:ffnn_in_out}) as shown in Algorithm
\ref{algo:ffnn_training} to complete an epoch. Generally the update happen over several epochs.

%% file: dia_ffnn_421.tex
\ifx\du\undefined
  \newlength{\du}
\fi
\setlength{\du}{12\unitlength}
\begin{tikzpicture}
\pgftransformxscale{1.000000}
\pgftransformyscale{-1.000000}
\definecolor{dialinecolor}{rgb}{0.000000, 0.000000, 0.000000}
\pgfsetstrokecolor{dialinecolor}
\definecolor{dialinecolor}{rgb}{1.000000, 1.000000, 1.000000}
\pgfsetfillcolor{dialinecolor}
\definecolor{dialinecolor}{rgb}{1.000000, 1.000000, 1.000000}
\pgfsetfillcolor{dialinecolor}
\pgfpathellipse{\pgfpoint{6.367911\du}{7.884825\du}}{\pgfpoint{0.927911\du}{0\du}}{\pgfpoint{0\du}{0.874825\du}}
\pgfusepath{fill}
\pgfsetlinewidth{0.0100000\du}
\pgfsetdash{}{0pt}
\pgfsetdash{}{0pt}
\pgfsetmiterjoin
\definecolor{dialinecolor}{rgb}{0.000000, 0.000000, 0.000000}
\pgfsetstrokecolor{dialinecolor}
\pgfpathellipse{\pgfpoint{6.367911\du}{7.884825\du}}{\pgfpoint{0.927911\du}{0\du}}{\pgfpoint{0\du}{0.874825\du}}
\pgfusepath{stroke}
\definecolor{dialinecolor}{rgb}{0.000000, 0.000000, 0.000000}
\pgfsetstrokecolor{dialinecolor}
\node at (6.367911\du,8.079825\du){};
\definecolor{dialinecolor}{rgb}{1.000000, 1.000000, 1.000000}
\pgfsetfillcolor{dialinecolor}
\pgfpathellipse{\pgfpoint{9.917911\du}{7.884825\du}}{\pgfpoint{0.927911\du}{0\du}}{\pgfpoint{0\du}{0.874825\du}}
\pgfusepath{fill}
\pgfsetlinewidth{0.0100000\du}
\pgfsetdash{}{0pt}
\pgfsetdash{}{0pt}
\pgfsetmiterjoin
\definecolor{dialinecolor}{rgb}{0.000000, 0.000000, 0.000000}
\pgfsetstrokecolor{dialinecolor}
\pgfpathellipse{\pgfpoint{9.917911\du}{7.884825\du}}{\pgfpoint{0.927911\du}{0\du}}{\pgfpoint{0\du}{0.874825\du}}
\pgfusepath{stroke}
\definecolor{dialinecolor}{rgb}{0.000000, 0.000000, 0.000000}
\pgfsetstrokecolor{dialinecolor}
\node at (9.917911\du,8.079825\du){};
\pgfsetlinewidth{0.0100000\du}
\pgfsetdash{}{0pt}
\pgfsetdash{}{0pt}
\pgfsetbuttcap
{
\definecolor{dialinecolor}{rgb}{0.000000, 0.000000, 0.000000}
\pgfsetfillcolor{dialinecolor}
\pgfsetarrowsend{stealth}
\definecolor{dialinecolor}{rgb}{0.000000, 0.000000, 0.000000}
\pgfsetstrokecolor{dialinecolor}
\draw (4.000000\du,12.050000\du)--(5.711779\du,8.503419\du);
}
\pgfsetlinewidth{0.0100000\du}
\pgfsetdash{}{0pt}
\pgfsetdash{}{0pt}
\pgfsetbuttcap
{
\definecolor{dialinecolor}{rgb}{0.000000, 0.000000, 0.000000}
\pgfsetfillcolor{dialinecolor}
\pgfsetarrowsend{stealth}
\definecolor{dialinecolor}{rgb}{0.000000, 0.000000, 0.000000}
\pgfsetstrokecolor{dialinecolor}
\draw (4.000000\du,11.900000\du)--(9.311779\du,8.703419\du);
}
\pgfsetlinewidth{0.0100000\du}
\pgfsetdash{}{0pt}
\pgfsetdash{}{0pt}
\pgfsetbuttcap
{
\definecolor{dialinecolor}{rgb}{0.000000, 0.000000, 0.000000}
\pgfsetfillcolor{dialinecolor}
\pgfsetarrowsend{stealth}
\definecolor{dialinecolor}{rgb}{0.000000, 0.000000, 0.000000}
\pgfsetstrokecolor{dialinecolor}
\draw (7.100000\du,11.850000\du)--(6.012815\du,8.693058\du);
}
\pgfsetlinewidth{0.0100000\du}
\pgfsetdash{}{0pt}
\pgfsetdash{}{0pt}
\pgfsetbuttcap
{
\definecolor{dialinecolor}{rgb}{0.000000, 0.000000, 0.000000}
\pgfsetfillcolor{dialinecolor}
\pgfsetarrowsend{stealth}
\definecolor{dialinecolor}{rgb}{0.000000, 0.000000, 0.000000}
\pgfsetstrokecolor{dialinecolor}
\draw (6.950000\du,11.900000\du)--(9.562815\du,8.693058\du);
}
\pgfsetlinewidth{0.0100000\du}
\pgfsetdash{}{0pt}
\pgfsetdash{}{0pt}
\pgfsetbuttcap
{
\definecolor{dialinecolor}{rgb}{0.000000, 0.000000, 0.000000}
\pgfsetfillcolor{dialinecolor}
\pgfsetarrowsend{stealth}
\definecolor{dialinecolor}{rgb}{0.000000, 0.000000, 0.000000}
\pgfsetstrokecolor{dialinecolor}
\draw (9.750000\du,11.900000\du)--(9.917911\du,8.759650\du);
}
\pgfsetlinewidth{0.0100000\du}
\pgfsetdash{}{0pt}
\pgfsetdash{}{0pt}
\pgfsetbuttcap
{
\definecolor{dialinecolor}{rgb}{0.000000, 0.000000, 0.000000}
\pgfsetfillcolor{dialinecolor}
\pgfsetarrowsend{stealth}
\definecolor{dialinecolor}{rgb}{0.000000, 0.000000, 0.000000}
\pgfsetstrokecolor{dialinecolor}
\draw (11.900000\du,12.000000\du)--(10.400000\du,8.800000\du);
}
\pgfsetlinewidth{0.0100000\du}
\pgfsetdash{}{0pt}
\pgfsetdash{}{0pt}
\pgfsetbuttcap
{
\definecolor{dialinecolor}{rgb}{0.000000, 0.000000, 0.000000}
\pgfsetfillcolor{dialinecolor}
\pgfsetarrowsend{stealth}
\definecolor{dialinecolor}{rgb}{0.000000, 0.000000, 0.000000}
\pgfsetstrokecolor{dialinecolor}
\draw (9.771314\du,11.876114\du)--(6.950000\du,8.700000\du);
}
\pgfsetlinewidth{0.0100000\du}
\pgfsetdash{}{0pt}
\pgfsetdash{}{0pt}
\pgfsetbuttcap
{
\definecolor{dialinecolor}{rgb}{0.000000, 0.000000, 0.000000}
\pgfsetfillcolor{dialinecolor}
\pgfsetarrowsend{stealth}
\definecolor{dialinecolor}{rgb}{0.000000, 0.000000, 0.000000}
\pgfsetstrokecolor{dialinecolor}
\draw (12.000000\du,12.100000\du)--(7.225189\du,8.219606\du);
}
\definecolor{dialinecolor}{rgb}{1.000000, 1.000000, 1.000000}
\pgfsetfillcolor{dialinecolor}
\pgfpathellipse{\pgfpoint{8.267911\du}{4.784825\du}}{\pgfpoint{0.927911\du}{0\du}}{\pgfpoint{0\du}{0.874825\du}}
\pgfusepath{fill}
\pgfsetlinewidth{0.0100000\du}
\pgfsetdash{}{0pt}
\pgfsetdash{}{0pt}
\pgfsetmiterjoin
\definecolor{dialinecolor}{rgb}{0.000000, 0.000000, 0.000000}
\pgfsetstrokecolor{dialinecolor}
\pgfpathellipse{\pgfpoint{8.267911\du}{4.784825\du}}{\pgfpoint{0.927911\du}{0\du}}{\pgfpoint{0\du}{0.874825\du}}
\pgfusepath{stroke}
\definecolor{dialinecolor}{rgb}{0.000000, 0.000000, 0.000000}
\pgfsetstrokecolor{dialinecolor}
\node at (8.267911\du,4.979825\du){};
\pgfsetlinewidth{0.0100000\du}
\pgfsetdash{}{0pt}
\pgfsetdash{}{0pt}
\pgfsetbuttcap
{
\definecolor{dialinecolor}{rgb}{0.000000, 0.000000, 0.000000}
\pgfsetfillcolor{dialinecolor}
\pgfsetarrowsend{stealth}
\definecolor{dialinecolor}{rgb}{0.000000, 0.000000, 0.000000}
\pgfsetstrokecolor{dialinecolor}
\draw (6.367911\du,7.010000\du)--(7.611779\du,5.403419\du);
}
\pgfsetlinewidth{0.0100000\du}
\pgfsetdash{}{0pt}
\pgfsetdash{}{0pt}
\pgfsetbuttcap
{
\definecolor{dialinecolor}{rgb}{0.000000, 0.000000, 0.000000}
\pgfsetfillcolor{dialinecolor}
\pgfsetarrowsend{stealth}
\definecolor{dialinecolor}{rgb}{0.000000, 0.000000, 0.000000}
\pgfsetstrokecolor{dialinecolor}
\draw (9.917911\du,7.010000\du)--(8.623007\du,5.593058\du);
}
\pgfsetlinewidth{0.0100000\du}
\pgfsetdash{}{0pt}
\pgfsetdash{}{0pt}
\pgfsetbuttcap
{
\definecolor{dialinecolor}{rgb}{0.000000, 0.000000, 0.000000}
\pgfsetfillcolor{dialinecolor}
\pgfsetarrowsend{stealth}
\definecolor{dialinecolor}{rgb}{0.000000, 0.000000, 0.000000}
\pgfsetstrokecolor{dialinecolor}
\draw (8.450000\du,3.850000\du)--(8.500000\du,2.050000\du);
}
\definecolor{dialinecolor}{rgb}{0.000000, 0.000000, 0.000000}
\pgfsetstrokecolor{dialinecolor}
\node[anchor=west] at (4.000000\du,12.650000\du){g1};
\definecolor{dialinecolor}{rgb}{0.000000, 0.000000, 0.000000}
\pgfsetstrokecolor{dialinecolor}
\node[anchor=west] at (6.900000\du,12.450000\du){g2};
\definecolor{dialinecolor}{rgb}{0.000000, 0.000000, 0.000000}
\pgfsetstrokecolor{dialinecolor}
\node[anchor=west] at (9.450000\du,12.400000\du){};
\definecolor{dialinecolor}{rgb}{0.000000, 0.000000, 0.000000}
\pgfsetstrokecolor{dialinecolor}
\node[anchor=west] at (9.500000\du,12.700000\du){g3};
\definecolor{dialinecolor}{rgb}{0.000000, 0.000000, 0.000000}
\pgfsetstrokecolor{dialinecolor}
\node[anchor=west] at (11.700000\du,12.650000\du){g4};
\definecolor{dialinecolor}{rgb}{0.000000, 0.000000, 0.000000}
\pgfsetstrokecolor{dialinecolor}
\node[anchor=west] at (9.000000\du,2.200000\du){o1};
\definecolor{dialinecolor}{rgb}{0.000000, 0.000000, 0.000000}
\pgfsetstrokecolor{dialinecolor}
\node[anchor=west] at (6.150000\du,6.350000\du){u1};
\definecolor{dialinecolor}{rgb}{0.000000, 0.000000, 0.000000}
\pgfsetstrokecolor{dialinecolor}
\node[anchor=west] at (9.650000\du,6.050000\du){};
\definecolor{dialinecolor}{rgb}{0.000000, 0.000000, 0.000000}
\pgfsetstrokecolor{dialinecolor}
\node[anchor=west] at (9.650000\du,6.000000\du){u2};
\definecolor{dialinecolor}{rgb}{0.000000, 0.000000, 0.000000}
\pgfsetstrokecolor{dialinecolor}
\node[anchor=west] at (4.050000\du,10.450000\du){w11};
\definecolor{dialinecolor}{rgb}{0.000000, 0.000000, 0.000000}
\pgfsetstrokecolor{dialinecolor}
\node[anchor=west] at (5.250000\du,11.100000\du){w21};
\definecolor{dialinecolor}{rgb}{0.000000, 0.000000, 0.000000}
\pgfsetstrokecolor{dialinecolor}
\node[anchor=west] at (11.250000\du,10.150000\du){w42};
\end{tikzpicture}

%% file: data_preperation.tex
 \section{Working scenario and Dataset preparation}
 \label{sec:dataset}

 \dataset\ 2016 dataset published for emotional impact of movies task 
is used in our analysis \cite{MediaEval2016}. 
This dataset is part of the LIRIS-ACCEDE dataset  \cite{LIRIS,7024148}
and consists of video 
clips of duration $8$-$12$ seconds which have been annotated by viewers for 
their perceived emotion, in terms of arousal and valance. Note that the 
perceived emotion annotation is for the entire video clip in terms of 
 valance and arousal value in the range $[0, 5]$. 

\begin{figure}
 \centerline{\includegraphics[width=0.27\textwidth,height=7.5cm]{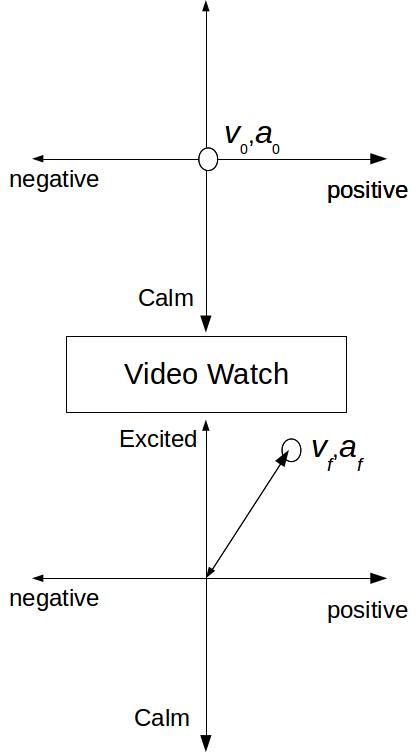}}
 \caption{Perceived Emotion after a viewer watches a short video.}
 \label{fig:dis_emo}
 \end{figure}

In this paper, to test our hypothesis, the problem of 
predicting the perceived valence (arousal) value of the viewer after watching 
the video is considered. Note that, as shown in Figure \ref{fig:dis_emo} 
the valence (arousal) value represent the emotional state of the viewer 
after having watched the video. It is not immediately clear if the perceived 
emotion annotated by the viewer is something that is perceived uniformly 
for the entire duration of the video clip or if the perceived emotion is based 
on a smaller segment which is the subset of the video clip.
According to the \cite{6890568,7024148}, each video clip in the dataset has a 
fade in at the beginning of the video clip and and a fade out at the end of the 
video clip. This implies that there is a priori knowledge in terms of how the 
emotion has a temporal relationship within the video  clip.
This aspect, namely the perceived emotion of a video clip has a fade in and fade out and is not uniform for the entire duration of the video clip 
motivates us to use \dataset\ 2016 dataset to evaluate 
our hypothesis. 

We created a dataset of smaller $1$ second video clips by segmenting 
the original video clip. Namely, a $10$ second original video clip produced 
$10$ $1$ second video clips, we retained the temporal relationship between the smaller video clips and the original video clip by naming the video clips 
appropriately. 
This enables us to incorporate the temporal correlations, in terms of the fade in and fade out, between the 
segments to test our hypothesis as mentioned in 
Section 
\ref{sec:hypothesis}. 

 In our experiments we concentrate only on the audio obtained from 
video clips as the input and the corresponding annotated valence (arousal) 
values are the desired output. For testing the hypothesis, 
we first extracted the 
audio from the original video clip and then segmented the audio into 
smaller non-overlapping $1$ second duration, so a movie clip 
of $n$ seconds ($ n \in [8, 12]$) duration, resulted in 
$n$ audio clips each of $1$ second duration. For example, if $c_k$ is 
the audio extracted from the original $k^{th}$ video then,
 \begin{equation}
 c_k = \oplus_{i=1}^{n} c_{ki}
 \label{eq:concatenation}
 \end{equation}
 where $\oplus$ represents the concatenation of the audio $c_{ki}$ for 
$i=1, \cdots, n$. Note that there is a temporal relationship between 
$c_{ki}$'s because they are in a time sequence and are from a single 
video clip. This construction (\ref{eq:concatenation}) helps us in 
building a dataset that can be used to analyze our hypothesis, namely, 
a FFNN infused with temporal knowledge can work as well as a 
RNN in terms of its overall performance when used for predicting the estimated emotion of a movie clip. 

\begin{figure}
 \centerline{\includegraphics[width=0.4\textwidth,height=4cm]{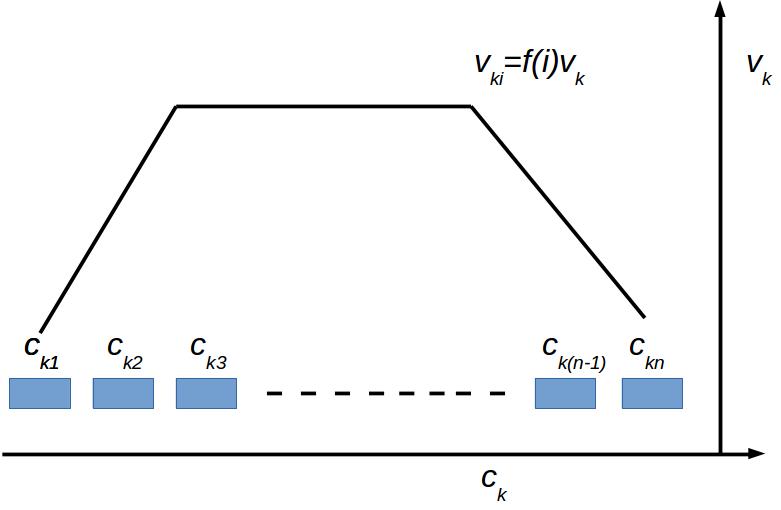}}
 \caption{$\{c_k; v_k\}$ pair}
 \label{fig:in_out}
 \end{figure}

 Let $\{c_k; o_k\}$ be the input output pair; where $o_k \in [0,5]$ can 
be either valence ($v_k$) or arousal ($a_k$) associated with the audio 
$c_k$. As seen in Figure \ref{fig:in_out}) the audio $c_k$ is made up of 
the $c_{k1}, c_{k2}, \cdots, c_{kn}$ audio sequence. So for a RNN we 
have the input as $c_{k1}, c_{k2}, \cdots, c_{kn}$ while the output is 
the associated $v_k$ (or $a_k$). However, since the input 
$c_{k1}, c_{k2}, \cdots, c_{kn}$ are temporally related, we assumed that 
the perceived valence $v_k$ (or 
arousal $a_k$)  has a bearing on $c_{ki}$. Namely,
 \begin{equation}
 v_{ki} = f(i) v_k
 \label{eq:in_out}
 \end{equation}
 For example, $f(i)$ could be a linear function,
 \[f(i) = \frac{i-1}{n-1}\]
 such that $v_{kn} = v_k$ and $v_{k1}=0$. Then each of the audio clips 
$c_{k1}, c_{k2}, \cdots, c_{kn}$ can be assigned a valence namely, 
$(c_{k1}; f(1)v_k)$, $(c_{k2}; f(2)v_k)$, $\cdots$, $(c_{kn}; f(n)v_k)$. 
Note that $f(i)$ captures the known a priori temporal knowledge.

We use $(c_k; v_k)$ or equivalently $(c_{k1}, c_{k2}, \cdots, c_{kn}; 
v_k)$ to train RNN while we use $(c_{k1}, \frac{0}{9}v_k)$, $(c_{k2}, 
\frac{1}{9}v_k)$, $\cdots$, $(c_{kn}, \frac{1}{1}v_k)$ to train a FFNN. 
Notice that for both FFNN and RNN the input data is the same while the 
output in case of RNN is known ($v_k$), we construct $v_{ki}$ using the 
prior knowledge for use in FFNN.

%% file: conclusions.tex
\section{Conclusions}
\label{sec:conclude}


FFNN architecture does not consider the temporal relationship that exits in a 
data sequence as in case of a speech signal. RNN architecture by its design 
is able to implicitly learn the temporal correlations that exists between the 
data sequence. While RNNs are advantageous when (a) one is not explicitly aware of the 
temporal relationship between the sequential data and (b) when there is a large
amount of training data. In this \paper, we address the scenario when there is 
insufficient training data and when a priori temporal knowledge about the 
training data is explicitly known.

In this \paper, we have shown how one can infuse explicitly known a priori 
temporal knowledge about the sequential data to enhance the performance of 
FFNN architecture. We first compared the differences between a simple RNN and 
a FFNN and showed that the a priori knowledge can in some sense compensate for 
the hidden layer feedback weights that contribute in capturing temporal 
relationship in the training data.  This observation, leads us to construct 
k-FFNN, a knowledge infused FFNN which exploits the known a priori sequential 
knowledge in the training data. This is one of the main contributions of 
this paper. Based on this observation, we hypothesized that k-FFNN performs 
as well as an RNN because k-FFNN are able to infuse known knowledge in the data sequence into its architecture. 
We further showed, experimentally, that the performance of k-FFNN 
especially for smaller training datasets exceeds the performance of RNN both in 
terms of the MSE and PCC and the performance of both k-FFNN and RNN level out 
when amount of training data increases. These experiments 
validate the hypothesis 
 {\em FFNN infused with prior knowledge about temporal relationship between data is similar to
RNN in terms of performance}.
The essential contribution of this \paper\ is the incorporation of known knowledge, when
available, to learn a FFNN without depending on a deep architecture like RNN that requires more samples for better training.